\documentclass[11pt]{article}

\usepackage[T1]{fontenc}
\usepackage{lmodern}
\usepackage[margin=1in]{geometry}
\usepackage{graphicx}
\usepackage{amsmath}
\usepackage{booktabs}
\usepackage{longtable}
\usepackage{array}
\usepackage{tabularx}
\usepackage{xcolor}
\usepackage{xurl}
\usepackage{hyperref}
\usepackage{caption}
\usepackage{enumitem}
\usepackage{parskip}
\usepackage{tikz}
\usepackage{float}
\usetikzlibrary{arrows.meta,positioning}

\hypersetup{
  colorlinks=true,
  linkcolor=blue,
  citecolor=blue,
  urlcolor=blue,
  pdftitle={Combining Language Models and Genetic Search for ARC-AGI-2},
  pdfauthor={Val Dyachenko}
}

\newcolumntype{Y}{>{\raggedright\arraybackslash}X}

\title{\textbf{Combining LLMs and Genetic Search for ARC-AGI-2}\thanks{Additional resources, including LLM conversations and GA runs, are available online:\par\noindent\mbox{\url{https://rockin.ai/blog/research/llm-seeded-genetic-search}}}}

\author{Val Dyachenko\\\texttt{valdytca@gmail.com}}
\date{September 2026}

\begin{document}
\maketitle

\begin{abstract}
LLMs can generate programs for ARC-AGI-2 tasks, but the provided compute only allows a small number of attempts to generate, debug and validate solutions. Genetic algorithms can search and test many more programs, but random search rarely starts in a useful neighborhood of the solution space. We combine the two methods through a compact domain specific language (DSL). First, a quantized Qwen3.5-4B LLM generates an initial set of programs for each ARC-AGI-2 task. Then, we use those programs to seed an initial population of starting programs, and use genetic algorithms to evolve these programs towards a solution to the given task. The DSL is designed such that every mutated program remains valid and can be executed. The initial programs proposed by the LLM solve 2 (3.3\%) of the first 60 tasks of the ARC-2 public evaluation set. The genetic algorithm solves an additional 4, giving 6 correct test outputs in total (10.0\%). If we try using evolving solutions without this LLM seeding, we do not arrive at any solutions at all. The results show that genetic search can improve programs generated by LLMs and produce additional correct solutions.
\end{abstract}

\section{Introduction}

ARC-AGI-2 is a benchmark with visual grid puzzles designed to test a system's abstract reasoning \cite{chollet2025arcagi2}. Each task provides a few examples that demonstrate how an input grid is transformed into an output grid. The system must discover the rule(s) and apply it to a new grid called the test input. The benchmark builds on the original Abstraction and Reasoning Corpus (ARC) \cite{chollet2019measure} and is normally quite easy for humans to solve, but extremely difficult for a machine due to the open ended nature of the tasks.

LLMs can generate programs for solving ARC-AGI-2 tasks, but limited time and compute gives them only a handful of attempts, which could be riddled with wrong reasoning and just plain bugs along the way. A genetic algorithm (GA) can test millions of programs, but a random seed has no knowledge which operations or parameters actually matter in the initial population. 

\textbf{Our main idea is to let the LLM suggest a useful starting point and let the GA search around it.} The LLM writes programs in a small domain-specific language (DSL). These programs then become the starting population for evolution. After this handoff, the LLM is no longer used.

We run every program on a small virtual machine, which we call the Machine. It has fixed grid color channels, registers, objects, and arguments. This structure is intentionally restrictive. Mutation and crossover could still produce a bad program, but they cannot produce broken syntax or a runtime error.

\begin{figure}[h]
\centering
\begin{tikzpicture}[
  box/.style={draw, rounded corners=2pt, align=center, text width=3.15cm, minimum height=1.25cm, fill=black!3, font=\small},
  flow/.style={-{Latex[length=2.2mm]}, semithick},
  label/.style={font=\scriptsize, fill=white, inner sep=1.5pt}
]
\node[box] (llm) {\textbf{LLM proposer}\\reason once; emit\\candidate programs};
\node[box, right=2.25cm of llm] (machine) {\textbf{Machine and DSL}\\parse, execute,\\compare with training outputs};
\node[box, right=2.25cm of machine] (ga) {\textbf{Genetic search}\\select, mutate,\\and recombine};

\draw[flow] ([yshift=5pt]machine.west) -- node[label, above] {state + DSL} ([yshift=5pt]llm.east);
\draw[flow] ([yshift=-5pt]llm.east) -- node[label, below] {seed programs} ([yshift=-5pt]machine.west);
\draw[flow] ([yshift=5pt]ga.west) -- node[label, above] {candidates} ([yshift=5pt]machine.east);
\draw[flow] ([yshift=-5pt]machine.east) -- node[label, below] {fitness} ([yshift=-5pt]ga.west);
\end{tikzpicture}
\caption{The LLM creates the initial programs. The GA then creates new programs, and the Machine executes and scores them. The LLM is not used during genetic search.}
\label{fig:system-overview}
\end{figure}
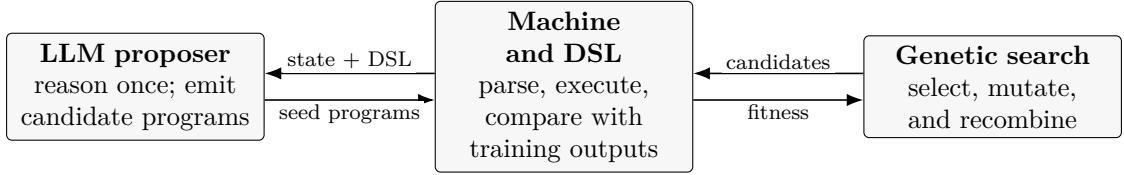

Figure~\ref{fig:system-overview} depicts how the high level system works. The LLM reads the Machine state for the task and operation list, then produces seed programs after getting a chance to reason about the solution approach (we use carefully written prompts to guide it). The GA runs the candidate programs through the Machine, and the Machine returns their fitness scores. This GA and Machine loop runs through generations until it finds a solution or times out.

We tested the method on 60 public ARC-AGI-2 evaluation tasks. The original LLM programs solve 2 tasks outright and after genetic search, an additional 4 are solved. In other words, the GA finds 4 correct programs that were missing from the original set of proposed solutions. If we let the GA run without LLM program seeds, it solves none of the 60 tasks. We tried different ways of randomly seeding the initial populations, but none of these attempts produced a working solution for any task. We therefore treat this finding as supporting evidence that the LLM is actually contributing real intelligence, albeit imperfect.

The paper makes three contributions:
\begin{itemize}[nosep]
  \item a transition from an LLM to a GA, with no further LLM calls during genetic search;
  \item a Machine and DSL that keep mutated programs valid and execute 8,000 candidate programs per generation on a GPU; and
  \item an overview that reports training fits and correct test outputs separately, including 2/60 task solves before genetic search and 6/60 after it.
\end{itemize}

\section{Related Work}

As of this writing, several ARC systems combine language models with program search. SOAR uses an LLM while running several rounds of evolutionary program creation and also fine-tunes the model between rounds \cite{soar2025}. HYSYNTH uses LLM completions to guide symbolic search \cite{hysynth2024}. ConceptSearch searches through LLM-generated ARC programs using scores based on task concepts \cite{conceptsearch2024}. ABPR solves ARC-AGI-2 tasks by asking an LLM to debug Prolog programs \cite{abpr2026}.

Our setup is different in one important way: the LLM is used only once before genetic search. It generates the initial programs and then leaves the loop. All later solutions are evolved by the GA with the fixed GPU executor. The programs also use a fixed-width and predefined DSL, so mutation and crossover cannot break their structure or create runtime errors. We leverage the LLM's ability to produce good but imperfect solutions, and the GA's ability to search over many similar candidates. 

Our implementation also takes advantage of running LLM conversations in parallel (with small context windows), and then running the genetic algorithms again in parallel on the same GPUs.

\section{The Technique}

\subsection{Overview}

The system goes through four steps for each ARC-2 task:
\begin{enumerate}[nosep]
  \item parse the train input grids into color channels, detected objects, and basic facts (i.e. x, y coordinates, width, height);
  \item prompt the LLM to think about the train inputs/outputs and generate DSL programs;
  \item use these programs to initialize two independent population islands and evolve them against the training examples; and
  \item run the first program that matches every training example on the test input.
\end{enumerate}

The GA search never gets to see the correct test output, and the programs are forced to generate the result in an output grid. A program is an \emph{exact training fit} if it reproduces every training output. We call a task \emph{solved} only if that same program also produces the correct test output.

\subsection{The Machine and program format}

The Machine is the shared state and language between the LLM and the GA. It stores a fixed set of grid color channels, up to eight detected objects, and 176 numeric registers. These registers hold useful precalculated facts such as grid size, colors, object centers, etc. A program, which we also call a chromosome, contains 20 genes. Each gene has an operation code and five arguments:
\begin{equation}
  g_i=[\mathrm{op},a,b,c,d,e], \qquad i\in\{1,\ldots,20\}.
\end{equation}

We designed this format specifically for the GA search, not for general programming. Every field has a fixed range, and every operation knows how to handle every allowed argument. Basically, every operation has its own implementation built into the Machine. Crossover swaps entire genes and mutation replaces fields with other allowed values. As a result, mutation cannot create a missing argument, an unknown variable, or invalid syntax. The new program may do the wrong thing, but the Machine can still run it. The GA search is also limited to 600 seconds per ARC-2 task, so expensive programs cannot use the whole competition budget (which is limited to 12 hours for 240 tasks).

These design decisions have a cost. A language with 41 operations can express far fewer ideas than plain Python. However, Python code would be near impossible to mutate safely and too slow to test at this scale. The LLM translates its ideas into the smaller language, and the GA searches through variations that the Machine can run quickly.

The program population uses three tensors: \emph{State} holds the grid channels, \emph{Registers} holds numeric values, and \emph{Genes} holds the programs. Table~\ref{tab:tensors} gives their shapes and contents.
\begin{table}[H]
\centering
\small
\begin{tabularx}{\linewidth}{@{}l l Y@{}}
\toprule
Tensor & Shape & Contents \\
\midrule
State & $[P,N,21,30,30]$ & Grid channels for $P$ programs and $N$ task examples, including OUT, color masks, objects, and scratch grids. \\
Registers & $[P,N,176]$ & Shared perceived facts plus per-program scratch registers. \\
Genes & $[P,20,6]$ & Twenty fixed-width genes per chromosome. \\
\bottomrule
\end{tabularx}
\caption{Machine tensors. Each island uses $P=4{,}000$ chromosomes.}
\label{tab:tensors}
\end{table}

\subsection{Preparing the ARC-2 task for the LLM}

Before calling the LLM, a deterministic perception step divides each grid into color masks and connected objects. It fills the object list and register bank. The LLM is given the same object and register names that the Machine uses. For example, when the LLM refers to the largest object's bounding box, the generated program can read and use that exact value from the registers.

\subsection{LLM proposal stage}
\label{sec:llm}

We use Qwen3.5-4B (quantized as Q4\_K\_M GGUF) and run it locally with \texttt{llama-server} from \texttt{llama.cpp} with thinking mode disabled. Each GPU runs its own server, and each conversation stays in one slot, so the llama server can reuse cached prompts. Before genetic search begins, each task goes through the same four LLM prompts:
\begin{enumerate}[nosep]
  \item \textbf{Plan:} give the train inputs and outputs, and parsed objects from perception, think about the solution, and express it in words;
  \item \textbf{Operations:} select correct operations from the list of available operations (all 41 of them);
  \item \textbf{Parameters:} generate parameters for those operations; and
  \item \textbf{Chromosomes:} generate complete programs in the fixed DSL syntax.
\end{enumerate}

Notice how we slowly guide the LLM towards building the programs. From our tests, this approach resulted in much better accuracy in the proposed solutions. The LLM first generates the likely rules/logic in plain text, and then chooses operations and arguments (and these become part of its context). This usually gives us several versions of the same basic idea instead of a single guess. The LLM generally goes in the right direction in its reasoning, but the exact details are not clear until it is asked to produce programs. We send the final chromosome request twice from the same conversation and allow 4,000 output tokens each time. We use only the valid and unique programs from both responses.

The LLM sometimes splits a multi-step solution into several neighboring programs. We keep each valid fragment as its own program and also join neighboring fragments in their original order, up to the 20-gene limit. This way, we test both the original proposals and the longer programs the LLM may have intended. For example, if a task requires several chained actions such as scale an object and then move it, the LLM could split these into separate programs.

\subsection{Genetic search}

We use two separate populations, or islands, with 4,000 programs each. Both islands are built the same. Forty percent of each island is seeded from the LLM programs. The other 60\% is random and uses the operations allowed for that task (referencing the original proposed programs). We found that having two islands gives us better chances at successfully solving a task.

For selection we use tournaments of size two. During crossover, each of the 20 gene positions is randomly copied from one of two parents. Each gene has a 0.09\% chance of mutation. If a gene mutates, each of its six parameters has a 0.5\% chance of being changed. The best 2\% of programs are promoted for the next generation unchanged (elitism strategy). The two islands do not exchange programs, in order to avoid both islands landing in the same local optimum. The search stops when either island finds the first exact training fit, reaches 200 generations, or runs over 600 seconds.

The fitness function is simple: every output pixel in each training pair has the same weight:
\begin{equation}
\mathrm{fitness}(p)=
\frac{\sum_{j=1}^{N}\sum_{(r,c)\in G_j}
\mathbf{1}\!\left[\hat{y}^{(p)}_{j,r,c}=y_{j,r,c}\right]}
{\sum_{j=1}^{N}|G_j|},
\end{equation}
where $G_j$ is the target grid for training pair $j$. A score of 1 means that the program perfectly reproduces every training output. We use the first program that reaches this score to produce the task output.

Table~\ref{tab:ga-config} lists the genetic search hyperparameters we used.
\begin{table}[H]
\centering
\small
\begin{tabular}{@{}lr@{}}
\toprule
GA parameter & Value \\
\midrule
Islands & 2 \\
Population per island & 4,000 \\
Chromosome length & 20 genes \\
LLM-derived / random initialization & 40\% / 60\% \\
Tournament size & 2 \\
Gene mutation probability & 0.09\%\\
Field resampling probability after mutation & 0.50\%\\
Elitism & 2\% \\
Migration & None \\
Termination & Exact fit, 200 generations, or 600 seconds \\
\bottomrule
\end{tabular}
\caption{Genetic-search parameters used in the reported run.}
\label{tab:ga-config}
\end{table}

\subsection{ARC-2 contest pipeline and compute}

We tuned this pipeline specifically for the ARC-AGI-2 contest and the time and computation provided by Kaggle. The pipeline first generates LLM programs for the tasks. This stage can use at most 60\% of the total runtime (12 hours). The seeded tasks then enter genetic search, one per available GPU. If time remains, the system runs more GA passes with new random states. The results are saved after every task run. This order ensures that slow LLM conversations cannot consume the time reserved for genetic search.

During development we used one consumer GPU with 4 GB of VRAM and Kaggle tests used two T4 GPUs. We also compiled \texttt{llama.cpp} for both T4 and L4 GPUs so the same packaged server could run in both Kaggle notebook environments.

\section{Task Evaluation and Test Results}

\subsection{Protocol}

We evaluate the first 60 tasks from the 120-task ARC-AGI-2 public evaluation set, ordered by task Id. The GA program search uses only the training examples. When it finds an exact training fit (a solution that correctly solves all training grids), we run that program on the test input and compare it with the published correct output. A task counts as solved only if the entire test grid is 100\% correct.

We compare the original LLM programs with the final programs found after genetic search on the same 60 tasks. We also ran control tests on the GA with random program seeds, and found no exact training fit, but this test configuration changed during development. For this reason, we do not treat it as a clean controlled experiment.

\subsection{Component outcomes}

Table~\ref{tab:main-result} compares the original LLM program pool with the result after genetic search.
\begin{table}[H]
\centering
\small
\begin{tabularx}{\linewidth}{@{}l r r Y@{}}
\toprule
Condition & Test solves & Solve rate & Interpretation \\
\midrule
LLM seed pool only & 2/60 & 3.3\% & Correct program already present before evolution. \\
LLM seed pool + GA & 6/60 & 10.0\% & Evolution adds four correct solutions! \\
\bottomrule
\end{tabularx}
\caption{Separating results from LLM and GA on the same 60 public-evaluation tasks.}
\label{tab:main-result}
\end{table}

The solve count increases from two to six. This shows that the GA actually does evolve programs towards a solution, and four additional correct programs emerge only after several generations of mutation and crossover. 

Among the 60 tasks, nine programs match each training sample. Six also produce the correct test output, while three fail on the test. Table~\ref{tab:exact-fits} lists all nine. Here, only the six correct test outputs count as solved tasks.

\begin{table}[H]
\centering
\small
\begin{tabular}{@{}lrrrl@{}}
\toprule
Task & Seed programs & Source & Generation & Test output \\
\midrule
\texttt{08ed6ac7} & 93 & LLM & -& Correct \\
\texttt{0b17323b} & 69 & LLM & -& Correct \\
\texttt{1190e5a7} & 46 & LLM & -& Incorrect \\
\midrule
\texttt{017c7c7b} & 73 & GA & 22 & Correct \\
\texttt{0c786b71} & 18 & GA & 35 & Correct \\
\texttt{0c9aba6e} & 45 & GA & 50 & Correct \\
\texttt{00576224} & 45 & GA & 80 & Correct \\
\texttt{0b148d64} & 156 & GA & 102 & Incorrect \\
\texttt{025d127b} & 59 & GA & 162 & Incorrect \\
\bottomrule
\end{tabular}
\caption{All exact training fits. Only rows marked ``Correct'' are counted as solved. }
\label{tab:exact-fits}
\end{table}

\subsection{Random seeded control}

We tried both the 27 core operations and the full 41-operation set during development and ran it on the same tasks that were solved by the LLM+GA combination. During these tests, the GA was unable to find any solutions.

Table~\ref{tab:ga-control} shows three example runs using only the core operations. We selected tasks that the combined system solves so that we could compare the search behavior directly. These examples are not the complete control dataset. Each run uses two unseeded populations for 200 generations.

\begin{table}[H]
\centering
\small
\begin{tabular}{@{}llr@{}}
\toprule
Task & Combined system & Unseeded GA best after 200 generations \\
\midrule
\texttt{0b17323b} & Exact at generation 0  & 79.73\% \\
\texttt{017c7c7b} & Exact at generation 22 & 93.50\% \\
\texttt{0c786b71} & Exact at generation 35 & 83.18\% \\
\bottomrule
\end{tabular}
\caption{Three illustrative exploratory runs with an unseeded, core-operation-only GA.}
\label{tab:ga-control}
\end{table}

\begin{figure}[t]
\centering
\includegraphics[width=0.72\linewidth]{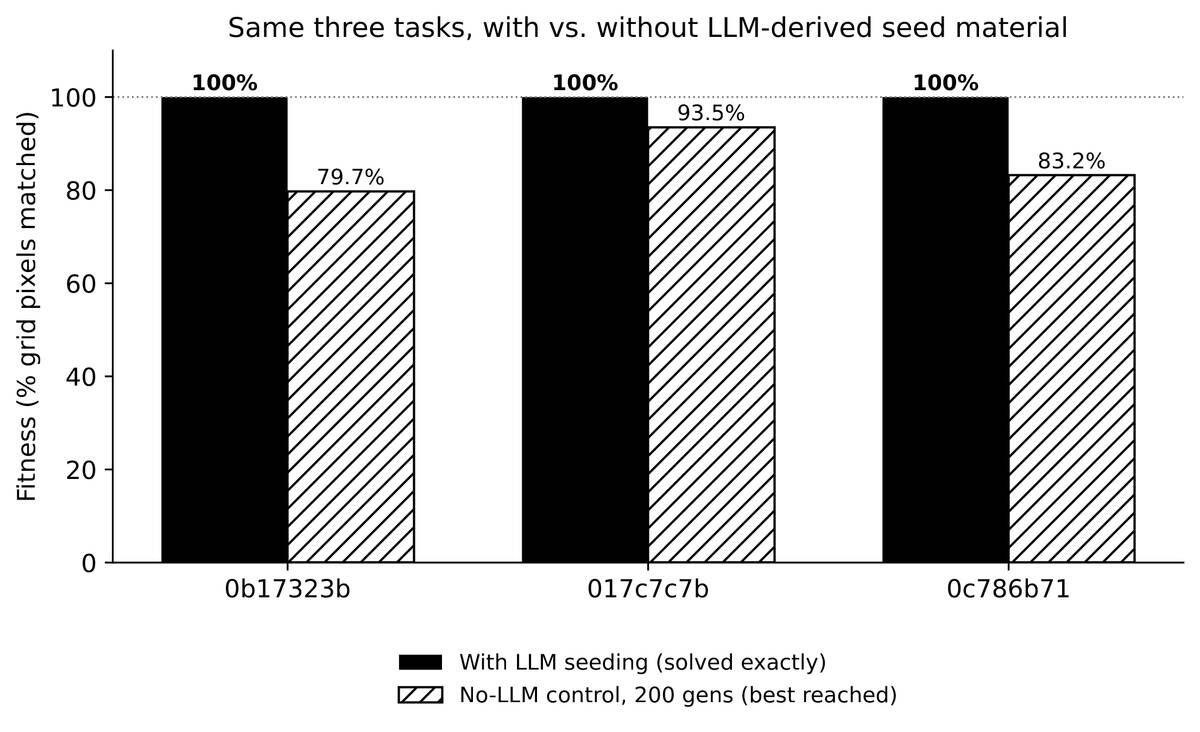}
\caption{Best fitness with LLM seeding and in three random seeded runs. Unfortunately, the random runs do not even reach an exact training task fit.}
\label{fig:ga-control}
\end{figure}

A stronger experiment in the future would keep all 41 operations and every other setting fixed, then repeat both seeded and unseeded runs on all 60 tasks, and potentially for longer runs than 200 generations.

\subsection{What evolution does}

The four additional correct programs appear after 22, 35, 50, and 80 generations. Some tasks improve through sudden jumps after several flat generations, and others improve through smaller steps. 

\begin{figure}[t]
\centering
\includegraphics[width=\linewidth]{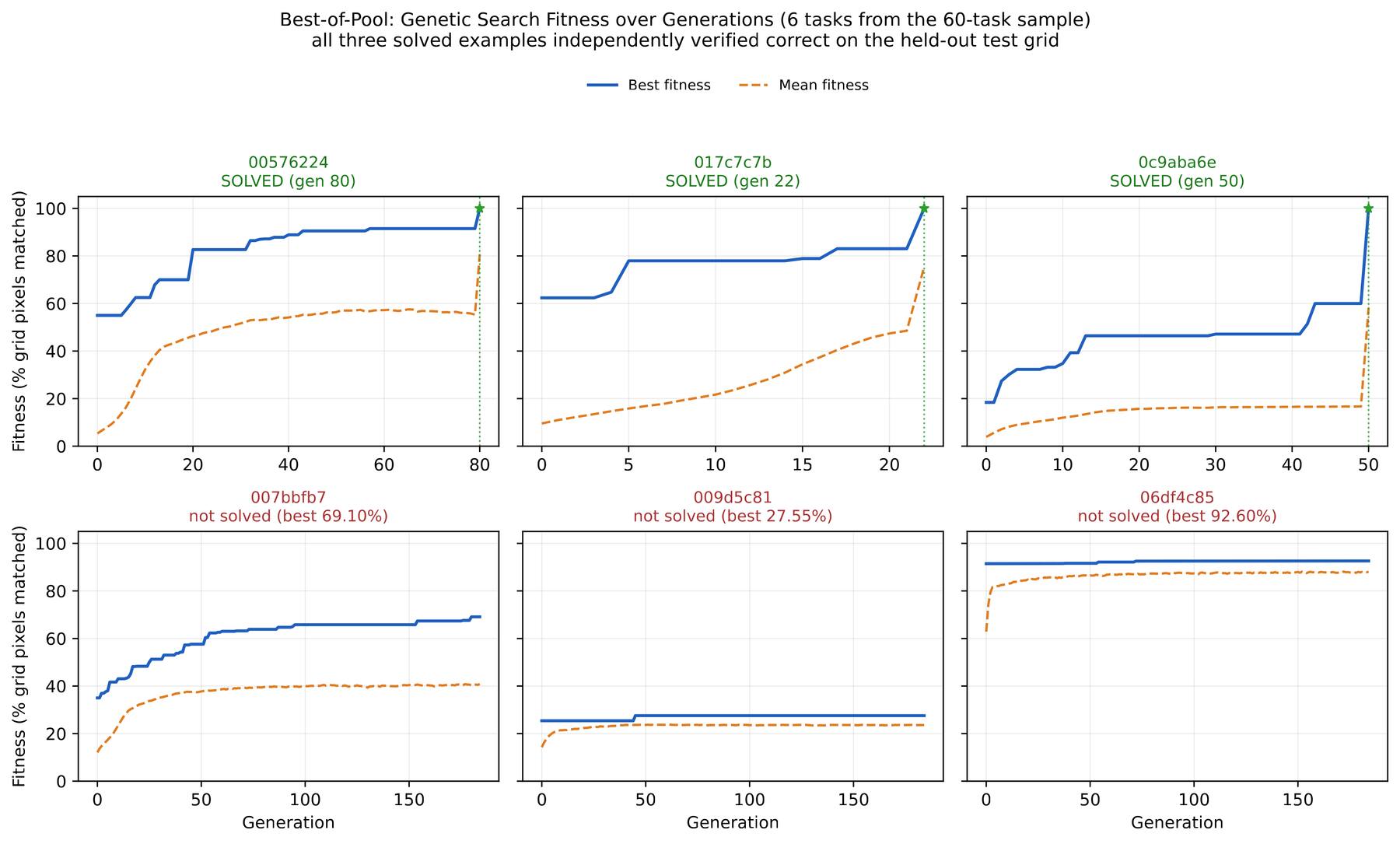}
\caption{Best and mean fitness for three solved and three unsolved tasks selected from the 60 task run. The examples illustrate the different convergence behaviors.}
\label{fig:convergence}
\end{figure}

The failed runs also show different problems. A low flat score likely means that the Machine simply cannot express the needed rule. A high flat score may mean that the population settled in a local optimum. A score that is still rising at the cutoff may simply need more generations; however, our runs are limited by resources.

\section{Discussion}

\subsection{Why the LLM+GA combination works}

The LLM and GA have different capabilities. The LLM is good at generating programs, choosing operations and arguments that appear relevant to the task. This gives the GA search a much better starting point than just random seeding.

The size of the DSL matters. If it is too small, some ARC-2 task solutions cannot be represented. If it is too general, mutations are more likely to not improve fitness. Our split between 27 core operations and 14 gated operations is one practical choice where common operations are always available and specialized operations become available only when the LLM uses them.

The six solved tasks do not show that this design is optimal; we still have a long way to go. They show something narrower: an LLM can produce useful starting programs, and a separate genetic search can turn some of them into new, correct solutions.

\subsection{Future work and research questions}

The first question is whether there is a better DSL set for this kind of search. An ideal DSL would match the reasoning abilities already present in an LLM while remaining small and regular enough for a GA to search millions of programs. Our 41-operation DSL is one point in this design space.

A second question is how the Machine and its operations should be designed and described. Qwen3.5-4B was \textbf{not} trained or tuned specifically for our Machine, yet it could read the Machine and produce programs that sometimes solved tasks in one try (we consider this a miracle all by itself). All of this suggests that it is possible to design another Machine that is even easier for an LLM to use correctly.

The third question is how far the GPU implementation can scale. The current executor handles 8,000 programs in parallel, with 20 genes per program and fixed $30\times30$ grids. More efficient kernels or memory layouts could support larger states, longer programs, or more operations. 

\subsection{Limitations}

We use the first 60 tasks by task Id, not a random sample or the full public evaluation set. This was mainly due to time and resource constraints. We used the first 20 of these tasks during development, so some tuning for a part of the evaluation set is possible. Second, the main result is stochastic and other runs may solve the same tasks at very different generations. Third, the random seed control was not complete, but it did verify our basic premise that the LLM was doing useful work. We therefore cannot measure exactly \textbf{how much} of the improvement comes from LLM seeding, even though the evidence points to it. Finally, the perception code and the 41-operation DSL does not and cannot represent every possible ARC rule. A task may fail because perception missed an object, or because an operation was missing.

\section{Conclusion}

This paper asks a simple question: can genetic search improve a set of programs generated by an LLM? In our experiment and with the proposed technique, it can. 

The LLM gives us a useful starting point and the Machine keeps all programs valid, running them quickly. The GA can then test about 1.6 million program instances for one ARC-2 task without making another LLM call. The LLM and GA perform different jobs, and the shared DSL allows them to work together.

The result is still preliminary but we hope the technique itself can be useful for other researchers who want to run similar searches through program solution spaces in the future.

\paragraph{Additional resources.}
LLM conversations, GA runs, and other implementation details are available online:

\noindent\mbox{\url{https://rockin.ai/blog/research/llm-seeded-genetic-search}}

\bibliographystyle{plain}

\clearpage
\appendix
\section{LLM Inference Configuration}

Table~\ref{tab:llm-config} records the inference settings used to generate the program pool.

\begin{table}[H]
\centering
\small
\begin{tabularx}{\linewidth}{@{}lY@{}}
\toprule
Setting & Value \\
\midrule
Model & Qwen3.5-4B, Q4\_K\_M GGUF \\
Topology & One \texttt{llama-server} per GPU; \texttt{--split-mode none}; all layers on the selected GPU \\
Parallelism & 10 conversation slots per GPU in the production pipeline (12gb VRAM) \\
Context & 35,000 tokens per slot; 350,000-token total server context at 10 slots \\
KV cache & Q8\_0 keys and values; Flash Attention enabled \\
Decoding mode & Thinking disabled with reasoning budget 0 and chat-template \texttt{enable\_thinking=false} \\
Sampling & Temperature and top-$p$ are not explicitly set; server defaults are used \\
Output budgets & 8,500 tokens for planning; 4,000 tokens for each chromosome response \\
Batching & Batch size 2,048; micro-batch size 1,024 \\
Acceleration & \texttt{ngram-simple} speculative decoding; observed approximately 1.8$\times$ throughput on repetitive DSL completions \\
\bottomrule
\end{tabularx}
\caption{LLM inference configuration used in the reported run.}
\label{tab:llm-config}
\end{table}

\clearpage
\section{Full DSL Operation Table}
\label{app:dsl}

\scriptsize
\setlength{\tabcolsep}{3pt}
\renewcommand{\arraystretch}{0.86}
\begin{longtable}{@{}p{0.7cm}p{2.4cm}p{1.0cm}p{9.0cm}@{}}
\toprule
\# & Operation & Tier & Function \\
\midrule
\endhead
0 & \texttt{NOP} & core & No-Op; fills unused chromosome gene positions. \\
1 & \texttt{COPY} & core & Copy one channel (color or object) to another. \\
2 & \texttt{TRANSLATE} & core & Shift content by a row and column offset. \\
3 & \texttt{RAY\_STEP} & core & Extend content repeatedly in a given direction. \\
4 & \texttt{BOOL\_OR} & core & Union of two channel channel grids. \\
5 & \texttt{BOOL\_AND} & core & Intersection of two channel grids. \\
6 & \texttt{BOOL\_XOR} & core & XOR of two channel grids. \\
7 & \texttt{BOOL\_SUB} & core & Subtract one channel from another. \\
8 & \texttt{RECOLOR} & core & Set all content cells of a channel to one color. \\
9 & \texttt{OVERLAY\_K0} & core & Paste content onto OUT while retaining existing grid. \\
10 & \texttt{READ\_CR} & core & Write a channel's centroid row to a register. \\
11 & \texttt{READ\_CC} & core & Write a channel's centroid column to a register. \\
12 & \texttt{PEEK\_COLOR} & core & Read the color at a grid coordinate into a register. \\
13 & \texttt{REG\_ADD} & core & Add two register values. \\
14 & \texttt{REG\_SUB} & core & Subtract two register values. \\
15 & \texttt{REG\_MUL} & core & Multiply two register values. \\
16 & \texttt{FLIP\_H} & core & Mirror left to right. \\
17 & \texttt{FLIP\_V} & core & Mirror top to bottom. \\
18 & \texttt{ROT180} & core & Rotate 180 degrees. \\
19 & \texttt{ROT90} & core & Rotate 90 degrees clockwise. \\
20 & \texttt{ROT270} & core & Rotate 90 degrees counter-clockwise. \\
21 & \texttt{TRANSPOSE} & core & Reflect across the main diagonal. \\
22 & \texttt{CROP\_BBOX} & core & Crop content to its bounding box. \\
23 & \texttt{TILE} & core & Repeat a block in a row-by-column grid. \\
24 & \texttt{MAP\_RANK} & core & Recolor objects according to their size rank. \\
25 & \texttt{PLACE} & core & Paste cropped content at an explicit coordinate. \\
26 & \texttt{EXTEND} & core & Extend content in a given direction to the grid boundary. \\
27 & \texttt{STRIPE} & gated & Extend a line ray with alternating colors. \\
28 & \texttt{CORNER\_RAY} & gated & Draw a diagonal ray from an object's empty bounding-box corner. \\
29 & \texttt{FILL\_BOX} & gated & Fill an axis-aligned rectangle on background cells. \\
30 & \texttt{DRAW\_LINE} & gated & Connect points with an axis-aligned bent path. \\
31 & \texttt{MOVE} & gated & Relocate cropped content and clear its previous footprint. \\
32 & \texttt{REG\_SELECT} & gated & Conditionally select a register value. \\
33 & \texttt{FLOOD} & gated & Fill enclosed background regions. \\
34 & \texttt{GRAVITY} & gated & Slide non-background cells until collision. \\
35 & \texttt{PICK} & gated & Select an object by a property such as size or border contact. \\
36 & \texttt{RECOLOR\_BY} & gated & Recolor objects according to a given rule. \\
37 & \texttt{PANEL\_BOOL} & gated & Split at a divider and Boolean-combine two panels in a grid. \\
38 & \texttt{SCALE} & gated & Scale up a given channel or object. \\
39 & \texttt{COPY\_PASTE} & gated & Periodically repeat content from its original position. \\
40 & \texttt{EXTEND\_ALL} & gated & Apply extension to every perceived object. \\
\bottomrule
\caption{The 41-operation DSL: 27 core operations and 14 LLM-gated operations.}
\end{longtable}

\end{document}